%% file: main.tex
\documentclass[letterpaper, 10 pt, conference]{ieeeconf}  

\IEEEoverridecommandlockouts                              

\usepackage{times} 
\usepackage{amsmath} 
\usepackage{amssymb}  
\usepackage{algpseudocode}
\usepackage[bookmarks=true]{hyperref}
\usepackage{array}
\usepackage{booktabs} 
\usepackage{graphicx}
\usepackage[utf8]{inputenc}
\usepackage{cite}
\usepackage{stfloats}
\usepackage{algorithm}
\title{\LARGE \bf ActGaze: Learning Action-Grounded Gaze through Counterfactual Visual Interventions for High-Precision Manipulation}

\author{
    Jinxuan Zhu\textsuperscript{1,2*},
    Jiaheng Wang\textsuperscript{1*},
    Chao Tang\textsuperscript{1},
    Mengfan Wang\textsuperscript{1},
    Hao Wei\textsuperscript{1},
    Shengbao Li\textsuperscript{1},
    \\
    Hong Yin\textsuperscript{1},
    Yiwen Gao\textsuperscript{2},
    Chenrui Tie\textsuperscript{2},
    and Tingguang Li\textsuperscript{1}\textsuperscript{$\dagger$}
    \\[0.5em]
    \small
    \textsuperscript{1}Samsung Robotics eXperience
    \quad
    \textsuperscript{2}National University of Singapore
    \\[0.3em]
    \small
    \textsuperscript{*}Equal contribution.
    \quad
    \textsuperscript{$\dagger$}Corresponding author.
}

\begin{document}

\maketitle
\thispagestyle{empty}
\pagestyle{empty}

\input{tex/abs}

\input{tex/intro}

\input{tex/related_work}

\input{tex/problem_formulation}
\input{tex/method}
\input{tex/exp}

\input{tex/conclusion}

\input{tex/Appendix}

\bibliographystyle{IEEEtran}
\bibliography{references}

\end{document}

%% file: tex/abs.tex
\begin{abstract}
Current Vision-Language-Action (VLA) models often struggle with high-precision robotic manipulation. We attribute this limitation primarily to their visual attention being dispersed across task-irrelevant regions. To address this issue, we propose \textbf{ActGaze}, a training approach that guides VLA policies to gaze on task-relevant regions, much like humans gaze on critical visual cues while executing precise movements. Unlike prior methods that rely on external labels for gaze supervision, ActGaze derives spatial supervision directly from the VLA's own action objective by using counterfactual visual interventions to identify regions that are critical for action prediction. Extensive real-robot experiments on four high-precision robotic manipulation tasks demonstrate that ActGaze induces more focused visual attention on task-relevant regions and consistently outperforms the base VLA policy and other visual-grounding approaches. More details can be found on: \href{https://anonymous.4open.science/w/ActGaze/}{ActGaze Homepage}.
\end{abstract}

%% file: tex/intro.tex
\section{Introduction}

Vision-Language-Action (VLA) models~\cite{black2024pi0,intelligence2025pi05,kim2024openvla,zitkovich2023rt2} have demonstrated strong generalization capabilities and broad potential
for robotic manipulation. However, current VLAs still struggle with
high-precision tasks such as insertion~\cite{zhu2025shapeforce} and assembly~\cite{tie2025manual2skill,tie2025manual2skill++}, which are
fundamental to many real-world and industrial applications.
Based on prior work~\cite{zhang2026embodied,song2026reconvla,vo2025clutter} and our
empirical analysis, we identify misallocated visual attention as a
major failure mode of VLA policies: rather than concentrating on
task-relevant regions, these models often attend to background clutter and visual distractors. This challenge is particularly pronounced in
high-precision manipulation, where only a small subset of visual regions
may contain information critical to the current action.
In standard end-to-end VLA training, the action objective supervises
the predicted actions but does not directly constrain where the model
should attend. As a result, task-relevant visual evidence may not be
explicitly prioritized during action prediction, potentially affecting
the robustness and generalization of VLA policies in high-precision  manipulation.

A natural way to address this limitation is to provide explicit
image-space supervision that guides the policy to gaze on visual
regions relevant to action generation. Existing approaches derive
such supervision from recorded human gaze~\cite{pani2026gaze,li2026gazevla,zuo2026gaze2act},
manually annotated masks or keypoints~\cite{sun2026artificial,murooka2026guidedattention},
and pseudo-labels generated by external models~\cite{song2026reconvla,lips2026generalization}.
While effective, these approaches require additional effort to obtain spatial supervision, limiting their widespread deployment. We therefore
seek to derive explicit spatial supervision directly from the action objective.

\begin{figure}[t]
    \centering
    \includegraphics[width=\linewidth]{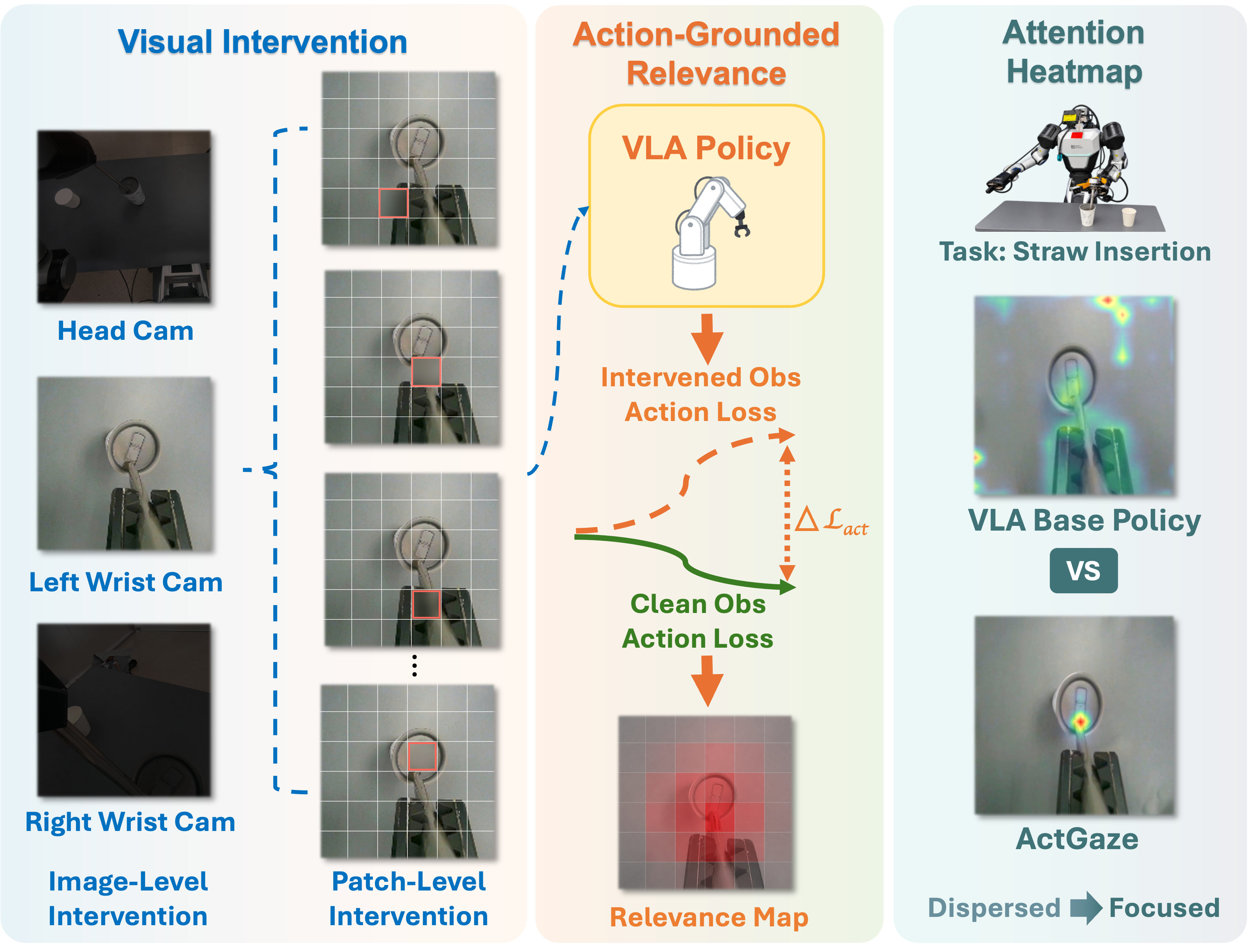 }
    \caption{\textbf{Overview of ActGaze}. ActGaze estimates the visual relevance via counterfactual visual intervention and action changes, guiding the VLA policy to gaze on task-relevant regions rather than dispersed visual cues.}
    \label{fig:teaser}
\end{figure}

To address the problem, we introduce \textbf{ActGaze}, an action-grounded gaze learning approach that derives explicit spatial supervision from the action objective through counterfactual visual interventions.
ActGaze builds on a simple counterfactual question
~\cite{pearl2009causal,zeiler2014visualizing,greydanus2018atari,zhang2026embodied,guo2024momentum}: \textit{how well would the policy predict the action if the visual evidence in a region were suppressed, while the rest of the observation remained unchanged?}
The resulting increase in action-prediction loss provides an estimate of the policy's dependence on that region. ActGaze turns these estimates into explicit spatial supervision to guide the policy's attention. Based on this idea, ActGaze first applies hierarchical counterfactual visual interventions to probe a VLA policy already trained on task demonstrations, measuring how interventions on different camera views and image regions affect its action predictions, as illustrated in Fig.~\ref{fig:teaser}. Image regions that cause larger degradation in action prediction are considered more important for decision-making, providing explicit spatial supervision for training a multi-view gaze head. The learned gaze is then transferred back to the VLA in the
gaze-to-action refinement stage, encouraging the VLA's action attention to align with the gaze head prediction. In this way, ActGaze converts intervention-induced changes in action prediction into explicit spatial guidance without requiring additional spatial annotations or external grounding models. 

We evaluate ActGaze on four real-world manipulation tasks requiring
sub-millimeter precision. ActGaze produces more task-relevant visual
attention and improves the average success rate from 77.0\% to 91.5\%. It also outperforms CoT
Grounding~\cite{zawalski2024ecot} and Implicit Grounding~\cite{song2026reconvla} in average success rate by 32.5\%
and 25.0\%, respectively, without requiring the external spatial supervision used by these baselines. These results demonstrate the effectiveness of action-grounded spatial guidance for high-precision manipulation.

We summarize our contributions as follows:

\begin{itemize}

    \item We introduce \textbf{ActGaze}, an action-grounded gaze learning approach that derives spatial supervision from intervention-induced changes in action prediction, without requiring external annotations or grounding models.

    \item We introduce counterfactual visual interventions to identify task-relevant visual regions, together with staged training that enables effective spatial knowledge distillation and policy refinement.

    \item We evaluate our method on four real-world manipulation tasks requiring sub-millimeter precision, demonstrating consistent improvements over the base policy and other visual-grounding approaches.

\end{itemize}

%% file: tex/related_work.tex
\section{Related Work}

\subsection{Spatial Guidance in VLA Policies}

Identifying task-relevant visual regions is important for action
generation in VLA policies.
In standard end-to-end training, the action objective supervises
predicted actions but does not directly constrain where the policy should attend. Consequently, VLA policies may attend to task-irrelevant
regions, limiting their ability to exploit fine-grained visual cues for high-precision manipulation. Existing approaches address this
limitation through two broad forms.

One line of work relies on external spatial supervision, including recorded human gaze~\cite{pani2026gaze,li2026gazevla,zuo2026gaze2act},
manually annotated masks or keypoints~\cite{sun2026artificial,murooka2026guidedattention},
and pseudo-labels generated by external models~\cite{pani2026gaze,qiu2026agra}. While effective, obtaining such supervision requires additional effort for data collection, annotation, or pseudo-label generation,
which can limit the widespread deployment of these approaches.

Another line derives spatial guidance from signals available within the policy or its execution history. These include execution
history~\cite{xiao2026ava}, spatial affinities encoded by the vision encoder~\cite{park2026ivra}, instruction-guided visual-token
pruning~\cite{li2026semanticvla}, adaptive visual-token
selection~\cite{zhang2026focusvla}, and attention signals associated
with action generation~\cite{liu2025vla,cui2026look}.
These approaches use internal representations, attention patterns,
or historical context as proxies for visual relevance. However,
such proxies may not directly reflect the policy's dependence
on a particular visual region for action prediction.

ActGaze uses hierarchical counterfactual visual interventions to
estimate this dependence from the resulting changes in action
prediction. These estimates provide explicit spatial supervision
without requiring additional spatial annotations or external
grounding models.

\subsection{Interventional Analysis of Action Policies}

Input perturbations have long been used to interpret vision models by modifying parts of an image and observing changes in model predictions~\cite{zeiler2014visualizing,fong2017interpretable,petsiuk1806rise,xing2023achieving}. This
paradigm has also been applied to sequential decision-making, where Greydanus et al.~\cite{greydanus2018atari} use perturbation-based saliency to analyze the visual behavior of learned policies. 

Recent studies use interventions to analyze and improve VLA policies and world-action models (WAM).
Embodied Interpretability~\cite{zhang2026embodied} formulates visual-action attribution as an interventional estimation problem and measures the influence of visual regions through changes in action prediction, while AGRA~\cite{qiu2026agra} uses causal interventions to reveal sensitivity to task-irrelevant visual regions. Related work also applies interventions to internal representations and multimodal pathways for interpretation, steering, and debiasing~\cite{haon2025mechanistic,shi2026vlatrace,
fang2026vision,zhang2026cofactvla}. In contrast, ActGaze goes beyond post hoc analysis by converting intervention-derived action relevance into spatial supervision for on-the-fly gaze learning, thereby improving VLA policy performance.

%% file: tex/problem_formulation.tex
\section{Problem Formulation}
\label{sec:problem}
We consider a VLA policy $\pi_{\theta}$ for robot manipulation. At each control step $t$, the policy takes multi-view RGB observations $\mathbf{I}_t=\{I_t^v\}_{v=1}^{V}$, the robot state $\mathbf{s}_t$ 
and a language instruction $\mathbf{l}$ as input, and predicts an $H$-step action chunk $\mathbf{A}_t=[\mathbf{a}_t,\ldots,\mathbf{a}_{t+H-1}]$ according to $\mathbf{A}_t \sim
\pi_{\theta}(\cdot\mid\mathbf{I}_t,\mathbf{s}_t,\mathbf{l})$.

However, action prediction alone does not explicitly specify which visual
regions are relevant for generating the action. We therefore aim to learn a
spatial gaze distribution that captures the action-relevant visual evidence
for the current task. Specifically, we introduce a gaze head
$G_{\phi}$ that estimates a relevance distribution over visual patches:
\begin{equation}
\hat{\mathbf{g}}_t
=
G_{\phi}
\left(
\mathbf{I}_t,\mathbf{s}_t,\mathbf{l}
\right),
\qquad
\sum_{v=1}^{V}\sum_{p=1}^{P_v}
\hat{g}_{t,v,p}=1,
\end{equation}
where $\hat{\mathbf{g}}_t$ denotes the relevance of patch
$p$ in camera view $v$. The joint normalization over views and patches allows
the gaze predictor to dynamically allocate visual relevance across different
camera views according to their task importance.

The goal of ActGaze is to learn an action-grounded gaze distribution
$\hat{\mathbf{g}}_t$ that identifies task-relevant visual regions and uses
this spatial guidance to improve VLA grounding for high-precision manipulation.

%% file: tex/method.tex
\section{Methodology}
\label{sec:method}

\begin{figure*}[t]
    \centering
    \includegraphics[width=\linewidth]{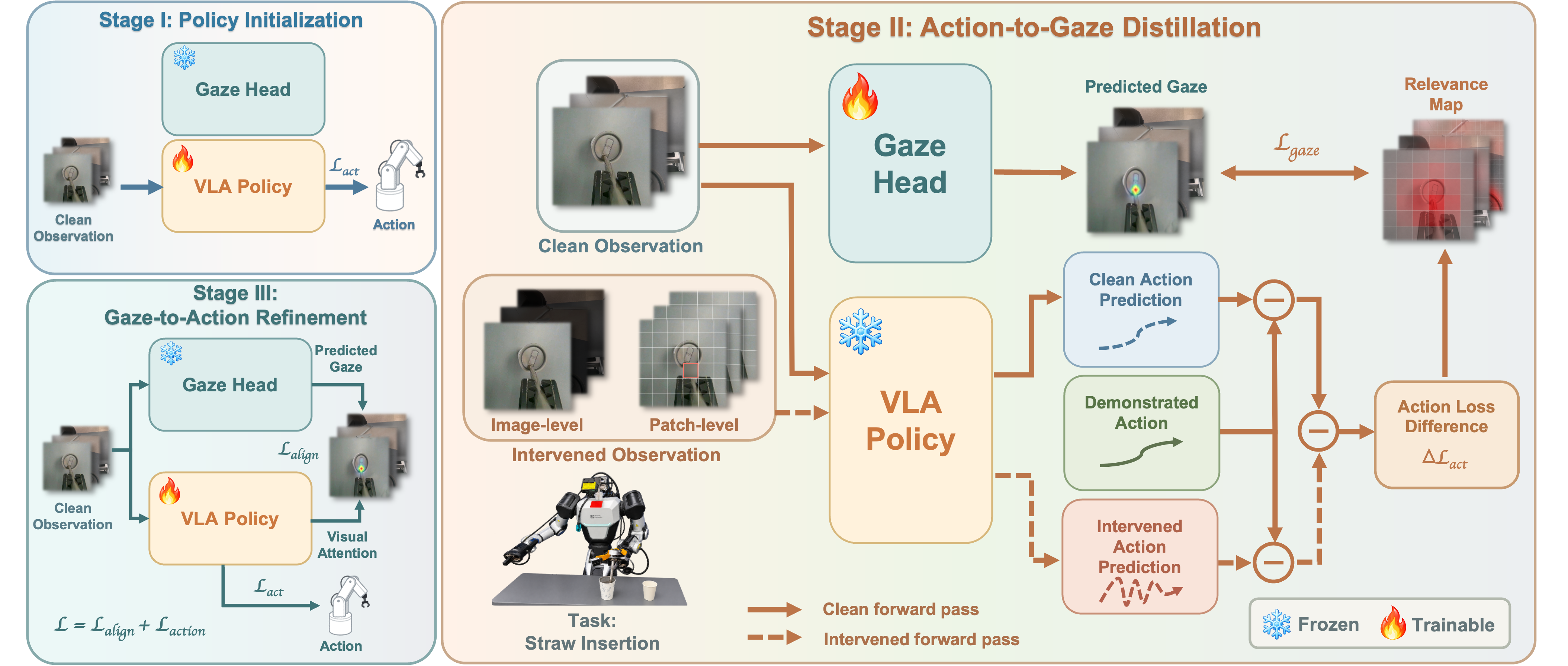}
    \caption{
    \textbf{Overview of ActGaze.}
    Stage I, \textbf{Policy Initialization}, trains the VLA until convergence to obtain
    an action-capable policy.
    Stage II, \textbf{Action-to-Gaze Distillation}, uses hierarchical
    counterfactual visual interventions to derive action-grounded spatial
    supervision and distill it into a multi-view gaze head.
    Stage III, \textbf{Gaze-to-Action Refinement}, transfers the learned gaze
    back to the VLA to refine its action-to-vision attention while preserving the primary action objective.
    }
    \label{fig:method}
\end{figure*}

ActGaze introduces a three-stage training approach for learning action-grounded gaze in VLA policies. In Sec.~\ref{sec:policy_init}, we describe
\textbf{Policy Initialization}, which establishes an action-capable VLA base policy for subsequent gaze learning by training on task demonstrations until convergence. Sec.~\ref{sec:action_to_gaze distillation} presents \textbf{Action-to-Gaze Distillation}, the core of ActGaze, which uses counterfactual visual interventions to derive action-grounded spatial targets and distills this knowledge into a multi-view gaze head. Finally,
Sec.~\ref{sec:policy_refinement} introduces
\textbf{Gaze-to-Action Refinement}, which transfers the learned gaze back to
the VLA to refine its action-to-vision attention while preserving the
primary action objective. Algo.~\ref{alg:ActGaze}  and Fig.~\ref{fig:method}  illustrate the overall
training procedure.

\begin{algorithm}[t]
\caption{\textbf{ActGaze}: Three-Stage Training Pipeline}
\label{alg:ActGaze}
\small
\begin{algorithmic}[1]

\Require Demonstrations $\mathcal{D}$, VLA policy $\pi_{\theta}$,
gaze head $G_{\phi}$
\Ensure Refined VLA policy $\pi_{\theta^*}$

\State \textbf{1. Policy Initialization}
\While{$\pi_{\theta}$ not converged}
    \State Update $\pi_{\theta}$ on $\mathcal{D}$ with
    $\mathcal{L}_{\mathrm{act}}$
\EndWhile
\State $\pi_{\theta} \rightarrow \pi_{\bar{\theta}}$

\State \textbf{2. Action-to-Gaze Distillation}
\State Freeze $\pi_{\bar{\theta}}$
\While{$G_{\phi}$ not converged}
    \State Sample observation $\mathbf{I}_t$
    \State Apply hierarchical visual intervention $\mathcal{I}(\mathbf{I}_t;S)$
    \State Compute action relevance $r_t(S)$
    \State Form gaze supervision from $r_t(S)$
    \State Update $G_{\phi}$ with $\mathcal{L}_{\mathrm{gaze}}$
\EndWhile
\State $G_{\phi} \rightarrow G_{\phi^*}$

\State \textbf{3. Gaze-to-Action Refinement}
\State Freeze $G_{\phi^*}$
\State Initialize $\pi_{\theta}$ from $\pi_{\bar{\theta}}$
\While{$\pi_{\theta}$ not converged}
    \State Predict gaze $\hat{\mathbf{g}}_t$ with $G_{\phi^*}$
    \State Extract attention $\boldsymbol{\alpha}_{\theta,t}^{v}$
    \State Update $\pi_{\theta}$ with $\mathcal{L}_{\mathrm{refine}}$
\EndWhile
\State $\pi_{\bar{\theta}} \rightarrow \pi_{\theta^*}$

\State \Return $\pi_{\theta^*}$

\end{algorithmic}
\end{algorithm}

\subsection{Policy Initialization}
\label{sec:policy_init}

Since ActGaze derives gaze supervision from the policy's action response,
we first train the VLA policy $\pi_{\theta}$ on action demonstrations
using its standard action-prediction objective $\mathcal{L}_{\mathrm{act}}$, so that it acquires a reliable action-prediction capability for action-to-gaze distillation in Sec.~\ref{sec:action_to_gaze distillation}.
We consider the policy sufficiently initialized when its validation
action loss falls below an empirical threshold $\epsilon_{\mathrm{act}}$.
Once this criterion is satisfied, we then freeze the policy as
$\pi_{\bar{\theta}}$, while the gaze head $G_{\phi}$ is not involved in this
stage.

It is worth noting that a converged action objective only ensures behavioral
competence but does not guarantee correct visual grounding. A policy may achieve
low action error by relying on correlated but non-essential visual cues rather
than task-relevant regions, leading to failures under small changes in object
pose, viewpoint, or visual distractions. Therefore, instead of directly
distilling the policy's existing attention patterns, we identify
action-relevant visual regions through counterfactual interventions on the
frozen policy.

\subsection{Action-to-Gaze Distillation}
\label{sec:action_to_gaze distillation}

Given the initialized policy from Sec.~\ref{sec:policy_init}, we freeze the VLA and train the gaze head $G_{\phi}$ from scratch. This stage consists of two steps. First, \textbf{Counterfactual Visual Intervention} measures how controlled perturbations of visual evidence affect action prediction, producing action-grounded relevance signals. Second, \textbf{Gaze Learning} distills these relevance signals into the gaze head, enabling it to predict
a task-relevant visual distribution from the original observations.

\subsubsection{Counterfactual Visual Intervention}

Given the frozen VLA policy, we aim to identify which visual regions are
important for its action prediction. We use \emph{counterfactual visual
intervention} 
to refer to a controlled intervention on a selected visual region while keeping the remaining inputs and action-prediction conditions unchanged. We then compare the policy's action prediction under the clean and intervened observations. If intervening on a visual region causes a larger degradation in action prediction, we regard that region as more relevant to the action, and vice versa. The resulting intervention-induced changes therefore provide an action-grounded supervisory signal for
gaze learning. Further theoretical analysis is provided in the Appendix.

Formally, given a visual region $S$, let
$\mathcal{I}(\mathbf{I}_t;S)$ denote the observation after intervening on
the visual evidence in $S$. We measure its action relevance as:
\begin{equation}
r_t(S)
=
\mathcal{L}_{\mathrm{act}}
\bigl(\mathcal{I}(\mathbf{I}_t;S)\bigr)
-
\mathcal{L}_{\mathrm{act}}(\mathbf{I}_t),
\label{eq:cf_relevance}
\end{equation}
where $\mathcal{L}_{\mathrm{act}}$ denotes the action-prediction loss
evaluated under the same non-visual inputs and action-prediction
conditions. A larger $r_t(S)$ indicates that intervening on $S$ causes
a larger degradation in action prediction, suggesting that the
corresponding visual evidence is more relevant to the action.

We apply hierarchical visual interventions $\mathcal{I}$ at two complementary spatial scales. At the camera level, $S$ spans an entire camera view, which is masked to estimate its relative relevance across views. At the patch level, $S$ corresponds to one of $K$ patch-aligned candidate regions within a view, which is blurred to estimate its local relevance to action prediction. The normalized camera-level relevance is then used to weight the patch-level gaze relevance.

\subsubsection{Gaze Learning}

The visual interventions above provide action-grounded relevance signals for the sampled regions $S$. We distill these sparse relevance
signals into a gaze head, enabling it to learn a coherent gaze distribution directly from the original observations while remaining computationally efficient.

For each camera view, the sampled $K$ candidate regions form a candidate set. We first convert their intervention responses $r_t(S)$ into a relative relevance distribution $\mathbf{q}_{t,v}$ by normalizing the responses
within this set. This relative normalization emphasizes the comparative importance among visual regions while reducing the influence of absolute
prediction changes caused by intervention-induced distribution shifts. The gaze head predicts patch-level logits from the clean observation, which
are aggregated over each candidate region and normalized to obtain
$\hat{\mathbf{q}}_{t,v}$. By aligning $\hat{\mathbf{q}}_{t,v}$ with $\mathbf{q}_{t,v}$, the gaze head learns to assign higher responses to regions that have a larger influence on action
prediction.

Since different camera views may contribute differently to action prediction, we weight the supervision from each view by its normalized camera-level relevance $w_{t,v}$. The gaze head is optimized using the weighted cross-entropy objective:
\begin{equation}
\mathcal{L}_{\mathrm{gaze}}
=
-\sum_{v=1}^{V} w_{t,v}
\sum_{k=1}^{K}
q_{t,v,k}
\log \hat{q}_{t,v,k},
\label{eq:gaze_learning}
\end{equation}
where $\sum_v w_{t,v}=1$. By sampling only a small set of candidate regions at each training step, ActGaze avoids exhaustive patch-wise interventions, improving computational efficiency while enabling on-the-fly generation of intervention-derived supervision. This distills sparse action-grounded relevance signals into $G_{\phi}$, enabling it to directly predict task-relevant visual attention
from the original observations. The learned gaze is then transferred back to the VLA to refine its visual attention and action prediction, as described in the next stage.

\subsection{Gaze-to-Action Refinement}
\label{sec:policy_refinement}

Finally, we freeze the learned gaze head and use its
predictions as spatial guidance to refine the VLA policy initialized from
$\pi_{\bar{\theta}}$. Let $\boldsymbol{\alpha}_{\theta,t}$ denote the action-to-vision attention
distribution of the VLA policy over all visual patches across camera views and $\hat{\mathbf{g}}_t$ denote the corresponding gaze prediction.
Before computing the refinement objective, both the action attention and gaze prediction are normalized over all valid visual patches across the multi-view observations.

We refine the VLA by jointly optimizing the original action objective and
a gaze-guided attention alignment term:
\begin{equation}
\mathcal{L}_{\mathrm{refine}}
=
\mathcal{L}_{\mathrm{act}}
+
\lambda_{\mathrm{align}}
\left\|
\boldsymbol{\alpha}_{\theta,t}
-
\hat{\mathbf{g}}_t
\right\|_2^2,
\label{eq:policy_refinement}
\end{equation}
where $\lambda_{\mathrm{align}}$ controls the strength of gaze-guided
attention alignment. The squared $L_2$ distance provides smooth spatial regularization, encouraging the policy attention
to follow the learned gaze while $\mathcal{L}_{\mathrm{act}}$ preserves its action-prediction
capability. This transfers
the learned spatial guidance back to the VLA and encourages more
consistent attention to task-relevant visual regions during action
prediction. Together, the three stages form a unified training pipeline that transitions seamlessly from policy initialization to gaze distillation and policy refinement within a single training run.

%% file: tex/exp.tex
\section{Experiment}
\label{sec:experiments}

Our experiments are designed to answer the following research questions:

\begin{enumerate}
    \item[\textbf{Q1.}]
    Can the gaze head learn \textbf{the correct task-relevant visual regions} from counterfactual visual interventions and effectively transfer them to the VLA policy? (Sec.~\ref{sec:Q1})

    \item[\textbf{Q2.}]
    Does ActGaze improve\textbf{ real-world high-precision manipulation} compared with the base VLA policy and existing visual grounding paradigms? (Sec.~\ref{sec:Q2})

    \item[\textbf{Q3.}]
    How do the \textbf{key design choices} of ActGaze contribute to its performance? (Sec.~\ref{sec:Q3})
    
\end{enumerate}

\begin{figure*}[t]
    \centering
    \includegraphics[width=0.95\linewidth]{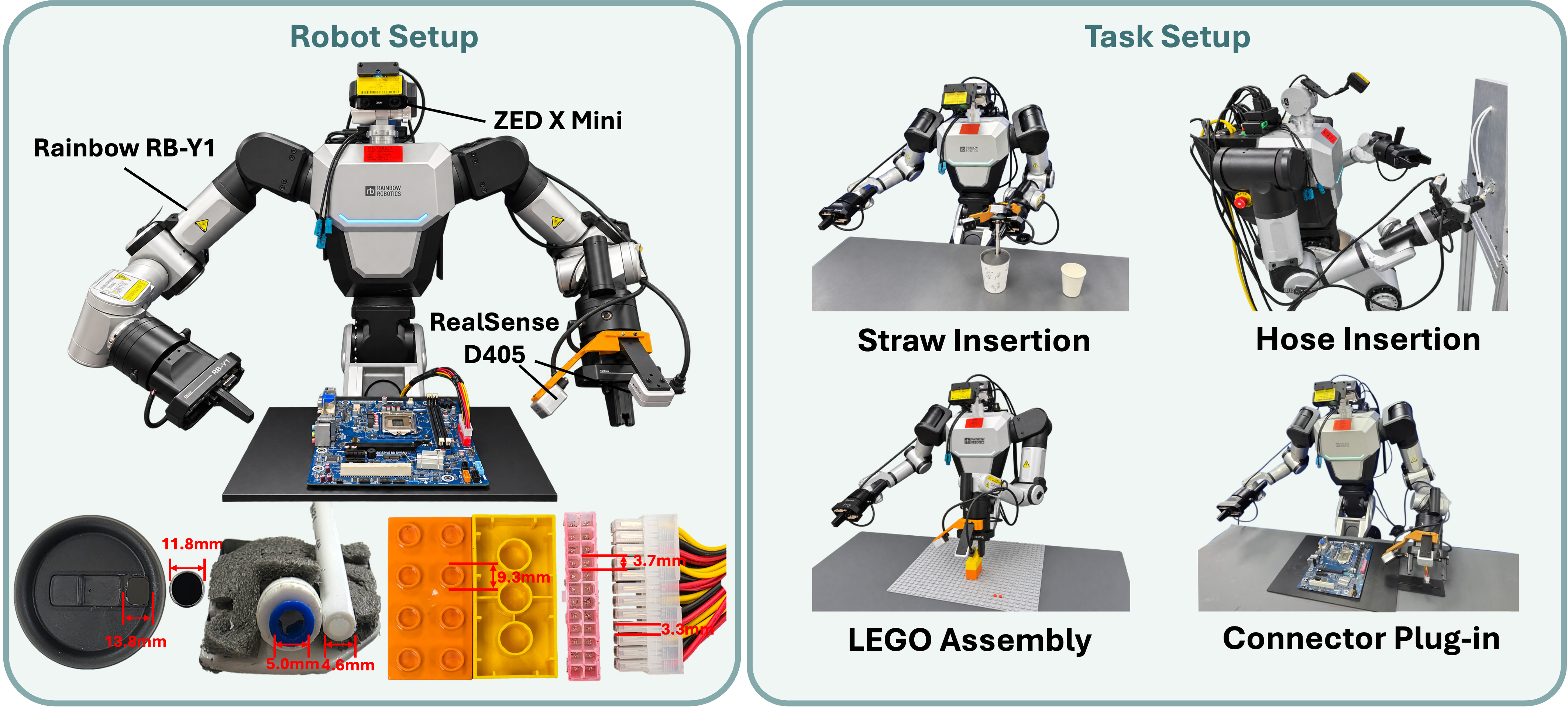}
    \caption{
    \textbf{Experimental Setup.}
    The left side illustrates the robot platform and object dimensions, while
    the right side presents the four precision manipulation tasks: Straw Insertion, Hose Insertion, LEGO Assembly and Connector Plug-in.
    }
    \label{fig:task_progress}
\end{figure*}
\subsection{Experimental Setup}

\textbf{Task Setup.} We evaluate ActGaze on four precision-demanding manipulation tasks:
Straw Insertion, Hose Insertion, LEGO Assembly, and Connector Plug-in,
as shown in Fig.~\ref{fig:task_progress}. These tasks cover diverse
fine-grained interactions, including rigid insertion, deformable object
manipulation, assembly, and multi-pin alignment. All tasks require
millimeter- to sub-millimeter-scale geometric precision, as illustrated by
the object dimensions in
Fig.~\ref{fig:task_progress}. We also introduce positional domain randomization to objects. The detailed
randomization ranges and task specifications are provided in Appendix. For brevity, we refer to these tasks as
\textit{Straw}, \textit{Hose}, \textit{LEGO}, and \textit{Connector} in the following tables.

\textbf{Implementation Details.}
We adopt $\pi_{0.5}$~\cite{intelligence2025pi05} as the base VLA policy and use a multi-layer
perceptron (MLP) as the gaze head $G_{\phi}$. For policy initialization,
we set the convergence threshold
$\epsilon_{\mathrm{act}}=0.002$. During gaze-to-action refinement in
Eq.~\ref{eq:policy_refinement}, we apply PCGrad~\cite{yu2020gradient}
to mitigate gradient conflicts between the action-prediction and gaze-alignment objectives. For computational efficiency, we set each candidate region introduced in Sec.~\ref{sec:action_to_gaze distillation} as a $2\times2$ block of four patches on the $16\times16$ visual patch grid and
sample only $K=2$ candidates per camera view at each training step. All policies are deployed on a Rainbow RB-Y1 robot. Visual observations are captured by two wrist-mounted Intel RealSense D405 cameras for local views and one head-mounted ZED X Mini camera for a global view of the
workspace. Policy training is conducted on eight NVIDIA RTX 5880 GPUs. For each task, all methods are trained using the same 200 expert demonstrations, batch size, and number of training steps to provide a fair
comparison. ActGaze and all baselines reported in the same table are also
evaluated within the same experimental session to reduce variations in
environmental and robot conditions.

\subsection{Action-Grounded Gaze Learning and Transfer (Q1)}
\label{sec:Q1}

\begin{table}[t]
\caption{
Task-relevant region alignment of the base policy attention, learned gaze, and refined policy attention, measured by  Ground-Truth Alignment Score $S_{\mathrm{GT}}$ (\%).
}
\label{tab:spatial_alignment}
\centering
\small
\resizebox{\linewidth}{!}{

\begin{tabular}{lccccc}
\toprule
Method & Straw & Hose & LEGO & Connector & Avg. ($\uparrow$) \\
\midrule
$\pi_{0.5}$ Action Attn.
& 14.0 & 12.0 & 21.2 & 17.9 & 16.3 \\
\textbf{ActGaze (Gaze Head)}
& \textbf{42.5} & \textbf{33.4} & \textbf{46.7} & \textbf{38.7} & \textbf{40.3} \\
\textbf{ActGaze (Action Attn.)}
& 37.3 & 30.7 & 38.0 & 36.9 & 35.7 \\
\bottomrule
\end{tabular}
}
\end{table}

\begin{figure}[t]
    \centering
    \includegraphics[width=\linewidth]{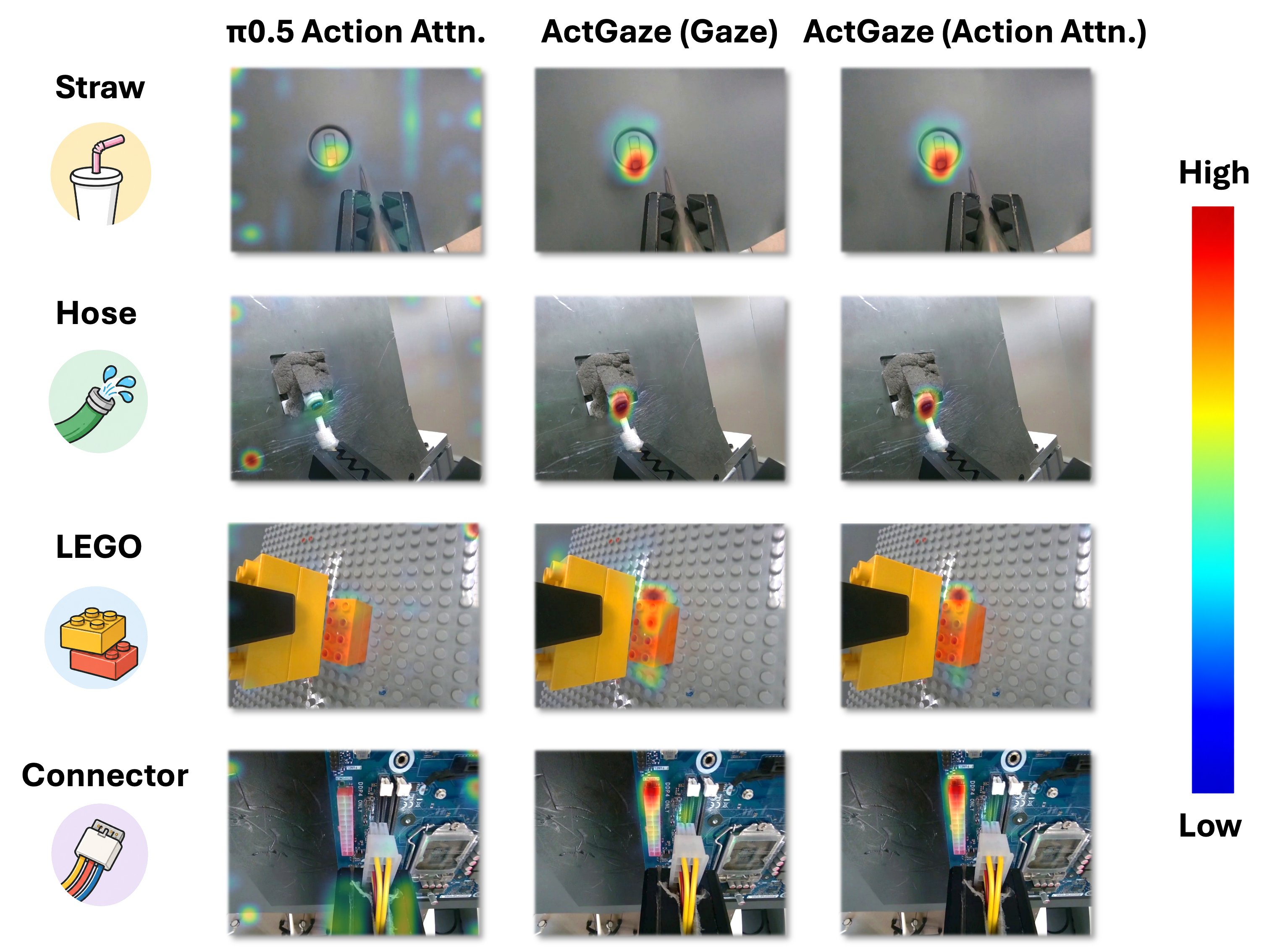}
    \caption{
    \textbf{Qualitative Comparison of Spatial Maps.} The learned gaze concentrates on task-relevant
    regions, and the refined policy attention shifts toward similar regions.
    }
    \label{fig:attention_heatmap}
\end{figure}

We first examine whether counterfactual visual intervention enables
ActGaze to identify task-relevant visual regions and whether the learned
spatial guidance can be effectively transferred back to the VLA policy.

\textbf{Evaluation Protocol.}
We manually annotate task-relevant regions with bounding boxes on 200 sparsely sampled frames
from each camera view of four tasks, collected across multiple episodes rather than from
consecutive video frames. These annotations are used solely as ground-truth regions for evaluation and are never exposed to ActGaze during training. The annotated regions cover the relevant grasping region and the target mating region. Examples can be found in the Appendix. We quantify spatial grounding using the Ground-Truth Alignment Score
$S_{\mathrm{GT}}$. The score measures the probability mass assigned to annotated task-relevant regions:
\begin{equation}
S_{\mathrm{GT}}(\mathbf{Z})
=
\frac{1}{N}
\sum_{t=1}^{N}
\sum_{v=1}^{V}
\sum_{p=1}^{P_v}
z_{t,v,p}m_{t,v,p},
\label{eq:gt_alignment}
\end{equation}
where $\mathbf{Z}=\{\mathbf{z}_t\}_{t=1}^{N}$ denotes the sequence of
normalized spatial maps, and $z_{t,v,p}$ represents the relevance assigned
to patch $p$ in camera view $v$ at time step $t$. The binary mask
$m_{t,v,p}\in\{0,1\}$ indicates whether the corresponding patch belongs to
the annotated task-relevant region.

We evaluate three spatial maps. \textbf{$\pi_{0.5}$ Action Attn.} denotes
the action-to-vision attention of the base VLA policy.
\textbf{ActGaze (Gaze Head)} corresponds to the gaze prediction
$\hat{\mathbf{g}}_t$ from gaze head, which evaluates whether
counterfactual supervision successfully identifies task-relevant regions.
\textbf{ActGaze (Action Attn.)} corresponds to
$\boldsymbol{\alpha}_{\theta,t}$ after gaze-to-action refinement, which
evaluates whether the learned spatial guidance is transferred back to the
VLA policy.

\textbf{Results.}
As shown in Table~\ref{tab:spatial_alignment}, the action attention of the
base $\pi_{0.5}$ policy achieves an average GT Alignment Score of only
16.3\%. In contrast, the learned gaze reaches 40.3\%, with consistent
improvements across all four tasks. This substantial improvement shows
that counterfactual visual intervention provides spatial information that
is substantially more aligned with task-relevant regions than the policy's
original attention. More importantly, after gaze-to-action refinement, the VLA's action
attention reaches 35.7\%, more than twice the alignment score of the base
policy. This demonstrates that the task-relevant spatial guidance learned
by the gaze head can be effectively transferred back to the VLA during
policy refinement. Figure~\ref{fig:attention_heatmap} provides qualitative results consistent
with these quantitative findings. Compared with the dispersed attention of
the base policy, the learned gaze concentrates more strongly on
task-relevant regions, while the refined policy attention shifts toward
similar regions. Overall, this experiment shows that ActGaze successfully learns task-relevant gaze from counterfactual visual interventions and transfers this spatial knowledge back to the VLA policy.

\subsection{Real-World High-Precision Manipulation (Q2)}
\label{sec:Q2}

We next evaluate whether the improved visual grounding learned by ActGaze translates into better real-world high-precision manipulation performance. We first compare ActGaze with the base $\pi_{0.5}$ policy on all four tasks,
and then compare different visual grounding paradigms that introduce auxiliary spatial supervision under the same $\pi_{0.5}$ backbone.

\begin{table}[t]
\caption{
Real-world success rates on four high-precision manipulation tasks comparing with base policy $\pi_{0.5}$.
}
\label{tab:main_success}
\centering
\small
\begin{tabular}{lccccc}
\toprule
Method
& Straw
& Hose
& LEGO
& Connector
& Avg. ($\uparrow$) \\
\midrule
$\pi_{0.5}$   & 37/50 & 34/50 & 44/50 & 39/50 & 77.0\% \\
\textbf{ActGaze} & \textbf{44/50} & \textbf{46/50} & \textbf{48/50} & \textbf{45/50} & \textbf{91.5\%} \\
\bottomrule
\end{tabular}
\end{table}
\textbf{Comparison with the Base Policy.} We first establish a controlled comparison with the base $\pi_{0.5}$ policy. Both methods use comparable numbers of policy optimization steps, and for each method we select the best checkpoint from a near-convergence window
using the same offline validation protocol. The selected policies are then evaluated under identical real-world conditions.

As shown in Table~\ref{tab:main_success}, ActGaze consistently improves success rates across all four high-precision tasks, increasing the average
performance from 77.0\% to 91.5\%. The largest improvement is observed on Hose insertion, where the success rate increases from 68.0\% to 92.0\%. We observe that this task contains strong specular reflections from the metal plate around the insertion opening, which can easily attract irrelevant visual attention. The larger gain on this task suggests that the action-grounded spatial guidance may help reduce sensitivity to such background distractions. We further examine this behavior under controlled visual distractors in the Appendix. These results demonstrate that the visual grounding improvements
observed in Sec.~\ref{sec:Q1} are not merely representational, but translate into substantial gains in real-world high-precision manipulation.

\begin{table}[t]
\caption{
Real-world success rates on two representative tasks comparing with other visual grounding paradigms.
}
\label{tab:grounding_comparison}
\centering
\small
\begin{tabular}{lccc}
\toprule
Method & Hose & Connector & Avg. ($\uparrow$) \\
\midrule
CoT Grounding
& 10/20 & 12/20 & 55.0\% \\
Implicit Grounding
& 13/20 & 12/20 & 62.5\% \\
\textbf{ActGaze}
& \textbf{18/20} & \textbf{17/20} & \textbf{87.5\%} \\
\bottomrule
\end{tabular}
\end{table}

\textbf{Comparison with Visual Grounding Paradigms.} To further evaluate the effectiveness of ActGaze in high-precision manipulation, we select two competitive visual grounding paradigms: CoT Grounding~\cite{zawalski2024ecot} and Implicit Grounding~\cite{song2026reconvla}.  
\begin{itemize}
    \item \textbf{CoT Grounding.}
    Following prior Chain-of-Thought grounding approaches ~\cite{zawalski2024ecot,deng2025graspvla}, CoT Grounding explicitly predicts the bounding-box coordinates of the task-relevant region as an intermediate grounding step prior to action sequence, encouraging the VLA to localize the
    target region before action prediction.

    \item \textbf{Implicit Grounding.}
    Following ReconVLA~\cite{song2026reconvla}, Implicit Grounding introduces an auxiliary reconstruction objective that reconstructs the task-relevant image crop during the training, encouraging the
    policy to implicitly encode where task-relevant visual information lies.
\end{itemize}

For a fair comparison, CoT Grounding, Implicit Grounding, and ActGaze all use the same VLA backbone and the same 200 expert demonstrations. The external bounding-box supervision required by CoT Grounding and Implicit Grounding is generated using Grounding-DINO~\cite{liu2023grounding}. 

As shown in Table~\ref{tab:grounding_comparison}, ActGaze achieves an average success rate of 87.5\%, outperforming CoT Grounding at 55.0\%
and Implicit Grounding at 62.5\% on both Hose Insertion and Connector Plug-in. Notably, the competing paradigms rely on external spatial
supervision, whereas ActGaze derives its spatial guidance directly from action demonstrations. 

Several factors may explain this difference. Spatial labels generated by external vision foundation models can contain localization errors or biases and may remain too coarse for high-precision manipulation, where critical cues such as openings, contact points, and mating interfaces often occupy only a small fraction of the grounded region. More fundamentally, high-precision manipulation often depends on the coupling between  interacting objects, and the relevant visual region changes with the manipulation phase~\cite{chen2026pointing}. The policy may need to attend to the manipulated object during pre-grasping, but shift toward the target opening or mating interface during alignment and insertion. External grounding mainly provides semantic object-level relevance, whereas ActGaze discovers relevance through its effect on the action objective. This makes the learned guidance naturally action-conditioned and phase-dependent, allowing it to focus on the visual evidence most critical to the current control step.

\subsection{Ablation Study of Key Design Choices (Q3)}
\label{sec:Q3}

Finally, we investigate the key design choices underlying ActGaze.
We first examine the role of counterfactual supervision in learning task-relevant gaze.
We then study the importance of staged training for stabilizing gaze learning.
Finally, we evaluate whether iterative refinement could provide significant gains beyond the first round.

\begin{table}[t]
\caption{
Ablation study of the key design choices.
}
\label{tab:component_ablation}
\centering
\small
\begin{tabular}{lccc}
\toprule
Method
& Straw
& LEGO
& Avg. ($\uparrow$)  \\
\midrule
w/o CF Supervision
& 8/20 & 7/20 & 37.5\% \\
w/o Staged Training
& 13/20 & 12/20 & 62.5\% \\
\textbf{ActGaze}
& \textbf{19/20} & \textbf{18/20} & \textbf{92.5\%} \\
\bottomrule
\end{tabular}
\end{table}

\textbf{Counterfactual Supervision.}  We preserve the same model architecture and
training stages as ActGaze, but remove the counterfactual supervision and
train the gaze head only through the action objective in the Stage~II in this ablation study. In this setting, the predicted gaze becomes nearly uniform over the visual
field, mainly because the converged action policy produces limited action loss variation with respect to visual changes, resulting in weak gradients for gaze head to learn spatial relevance. Moreover, aligning action attention with this nearly uniform target may damage the useful spatial knowledge already present in the
initialized policy, impairing action prediction. As a result, the average success rate drops to 37.5\%, compared with
92.5\% for ActGaze in the Straw Insertion and LEGO Assembly tasks as shown in Table~\ref{tab:component_ablation}. This suggests that the benefit of ActGaze comes
not from simply introducing a gaze head, but from explicitly converting action sensitivity into spatial supervision through counterfactual
interventions.

\textbf{Staged Training.}
We remove the three-stage training pipeline and jointly train the gaze head and VLA policy with both counterfactual supervision and the action objective from the beginning. We observe that the predicted gaze continuously drifts during training and does not consistently concentrate on task-relevant regions. This is because
the VLA policy has not yet learned to predict actions reliably during early training, resulting in unreliable counterfactual supervision for gaze learning. Accordingly, this variant achieves an average success rate of only 62.5\%, compared with 92.5\% for
ActGaze with staged training in Straw Insertion and LEGO Assembly tasks, as shown in Table~\ref{tab:component_ablation}. These results highlight the need
for VLA initialization to establish a stable supervisory signal for gaze learning. Although convergence in Stage~I does not necessarily guarantee strong real-world performance, it provides a stable action-prediction reference that is essential for subsequent gaze learning in our pipeline.

\begin{figure}[t]
    \centering
    \includegraphics[width=\linewidth]{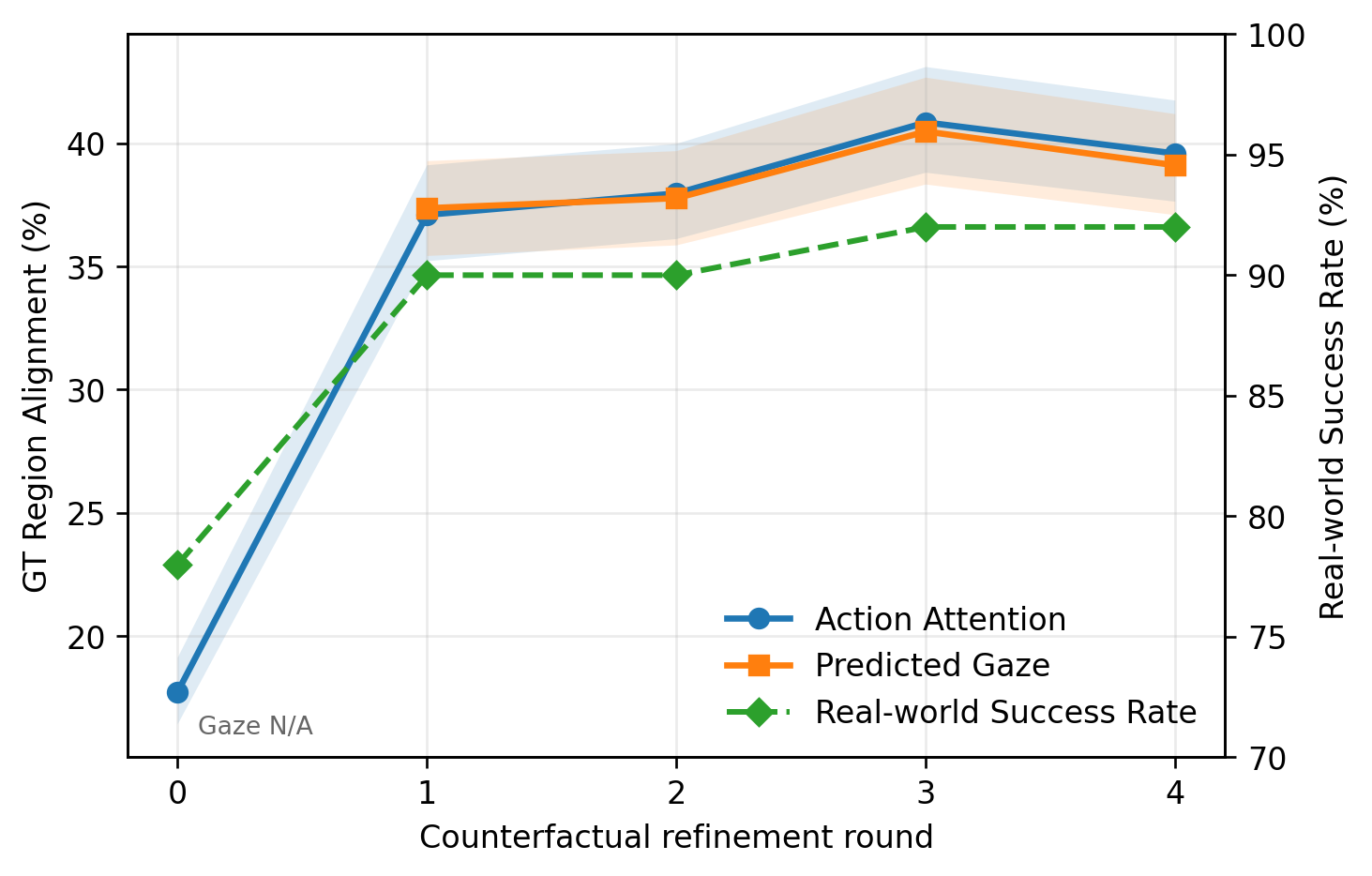}
    \caption{
   \textbf{ Iterative Refinement Result.} We plot $S_{\mathrm{GT}}$ and real-world success rate across multiple training rounds in LEGO Assembly. Most improvements are achieved in the first round, while later rounds yield only marginal improvements.
}
    \label{fig:iterative_refinement}
\end{figure}

\textbf{Iterative Refinement.}
We investigate whether iterative refinement provides further improvements. Starting from the policy obtained in the previous round, we repeat action-to-gaze distillation and gaze-to-action refinement for four rounds in LEGO Assembly task. We report $S_{\mathrm{GT}}$ for both predicted gaze and action attention, along with the real-world success rate. As shown in Fig.~\ref{fig:iterative_refinement}, the success rate increases from 78\% at Round~0 to 90\% at Round~1, remains at 90\%
at Round~2, and reaches 92\% at Rounds~3--4. Ground-Truth Alignment Score $S_{\mathrm{GT}}$ likewise improves substantially in the first round, with only marginal gains thereafter, consistent with the success-rate trend. This pattern suggests the gain from iterative refinement is not significant, which may result from the limitation of performance upper bound and bias from initial model. We therefore use one round of refinement to balance the gain and the computational cost.

%% file: tex/conclusion.tex
\section{Conclusion and Limitations}
\label{sec:conclusion}

We introduced \textbf{ActGaze}, an action-grounded gaze learning approach that converts intervention-induced changes in action prediction into explicit spatial supervision. Experiments on four real-world high-precision manipulation tasks demonstrate consistently improved visual attention and task performance without requiring external annotations or grounding models for supervision. The framework is also potentially applicable to a broad class of visuomotor policies.

However, ActGaze has several limitations. First, it performs counterfactual analysis at the visual level rather than establishing causal relationships in the physical environment. Intervention responses may also be affected by perturbation-induced distribution shifts and should therefore be interpreted as relevance estimates rather than strong causal effects. Second, because spatial supervision is derived from the initial policy, gaze learning remains constrained by its existing knowledge and biases. Such biases cannot be fully eliminated through distillation and may persist across refinement rounds, limiting further performance gains. Despite these limitations, the observed improvements support the practical effectiveness of ActGaze. We hope this work can inspire further exploration of self-evolving learning-based policies.

%% file: tex/Appendix.tex
\appendix

This appendix complements the main paper with additional theoretical
motivation, implementation details, and evaluation results for ActGaze.
We first explain why the action-loss change induced by counterfactual
visual interventions provides a principled signal for identifying
action-relevant regions (Sec.~\ref{app:cf_theory}). We then describe the
ground-truth region annotation protocol used in our evaluation
(Sec.~\ref{app:gt_annotation}) and provide detailed task specifications
and randomization settings to facilitate reproducibility
(Sec.~\ref{app:tasks}). Finally, we present additional experiments that
evaluate the robustness of ActGaze under visual distractors
(Sec.~\ref{app:distractors}).

\subsection{Theoretical Analysis of Counterfactual Visual Intervention}
\label{app:cf_theory}
The main paper introduces counterfactual visual intervention as a practical
way to estimate which visual regions are relevant to action prediction.
Here, we provide additional theoretical motivation for this design by
connecting intervention-induced loss changes to the predictive value of
visual evidence.
\paragraph{Predictive value under ideal information removal.}
Let $X$ denote the complete visual observation, $X^{-S}$ the visual
observation with the information in region $S$ removed, $C$ all remaining
prediction conditions, and $Y$ the prediction target.

Under squared-error loss, the optimal predictor with access to the full
visual observation is
\begin{equation}
\hat{Y}
=
\mathbb{E}[Y \mid X,C],
\end{equation}
while the optimal predictor without access to region $S$ is
\begin{equation}
\hat{Y}^{-S}
=
\mathbb{E}[Y \mid X^{-S},C].
\end{equation}

We define the predictive value of region $S$ as the increase in expected
prediction loss when the information in $S$ is unavailable:
\begin{equation}
\mathcal{V}(S)
=
\mathbb{E}\!\left[
\|Y-\hat{Y}^{-S}\|_2^2
\right]
-
\mathbb{E}\!\left[
\|Y-\hat{Y}\|_2^2
\right].
\label{eq:app_predictive_value_def}
\end{equation}

Intuitively, $\mathcal{V}(S)$ measures how much prediction performance is
lost when the visual evidence in $S$ is removed.

\textbf{Proposition 1.}
The predictive value defined above satisfies
\begin{equation}
\mathcal{V}(S)
=
\mathbb{E}\!\left[
\|\hat{Y}-\hat{Y}^{-S}\|_2^2
\right]
\geq 0.
\label{eq:app_predictive_value}
\end{equation}

\textit{Proof.}
We first decompose the prediction error without region $S$ as
\begin{equation}
Y-\hat{Y}^{-S}
=
(Y-\hat{Y})
+
(\hat{Y}-\hat{Y}^{-S}).
\end{equation}
Taking the squared norm and expectation gives
\begin{equation}
\begin{aligned}
\mathbb{E}\!\left[
\|Y-\hat{Y}^{-S}\|_2^2
\right]
=&~
\mathbb{E}\!\left[
\|Y-\hat{Y}\|_2^2
\right] \\
&+
\mathbb{E}\!\left[
\|\hat{Y}-\hat{Y}^{-S}\|_2^2
\right] \\
&+
2\mathbb{E}\!\left[
(Y-\hat{Y})^\top
(\hat{Y}-\hat{Y}^{-S})
\right].
\end{aligned}
\end{equation}

The last term is zero. Since $\hat{Y}-\hat{Y}^{-S}$ is determined by the
observed inputs $X$ and $C$,
\begin{equation}
\begin{aligned}
&\mathbb{E}\!\left[
(Y-\hat{Y})^\top
(\hat{Y}-\hat{Y}^{-S})
\right] \\
&=
\mathbb{E}\!\left[
(\hat{Y}-\hat{Y}^{-S})^\top
\mathbb{E}[Y-\hat{Y}\mid X,C]
\right]
=0,
\end{aligned}
\end{equation}
because
\begin{equation}
\mathbb{E}[Y-\hat{Y}\mid X,C]=0.
\end{equation}

Therefore,
\begin{equation}
\mathbb{E}\!\left[
\|Y-\hat{Y}^{-S}\|_2^2
\right]
=
\mathbb{E}\!\left[
\|Y-\hat{Y}\|_2^2
\right]
+
\mathbb{E}\!\left[
\|\hat{Y}-\hat{Y}^{-S}\|_2^2
\right].
\end{equation}
Substituting this result into
Eq.~\ref{eq:app_predictive_value_def} yields
Eq.~\ref{eq:app_predictive_value}.
Since a squared norm is always non-negative,
$\mathcal{V}(S)\geq 0$.

Equation~\ref{eq:app_predictive_value} provides a direct interpretation of
visual relevance. If removing region $S$ does not change the optimal
prediction, then $\hat{Y}=\hat{Y}^{-S}$ and $\mathcal{V}(S)=0$. In contrast,
if the information in $S$ substantially affects the prediction, the two
predictors differ more strongly and $\mathcal{V}(S)$ becomes larger.

Importantly, this predictive value is context-dependent. Information from
other visual regions or camera views may compensate for the removal of $S$.
Therefore, the relevance of a region is defined relative to the remaining
visual context, rather than being an intrinsic property of the region alone.

\paragraph{From ideal information removal to ActGaze.}
The ideal quantity $\mathcal{V}(S)$ requires prediction under the true
conditional distribution with region $S$ unavailable, which is generally
intractable for a learned VLA policy. ActGaze therefore uses a practical
single-policy approximation. Let $\mathcal{I}(I;S)$ denote the observation
obtained by suppressing the visual evidence in $S$. We compute
\begin{equation}
r(S)
=
\mathcal{L}_{\mathrm{act}}
\left(
\mathcal{I}(I;S)
\right)
-
\mathcal{L}_{\mathrm{act}}(I),
\label{eq:app_intervention_score}
\end{equation}
where the initialized policy is kept fixed.

This construction approximates the ideal comparison in two ways.
First, the initialized policy serves as an approximation to the
full-information predictor. Second, visual intervention approximates the
prediction that would be obtained if the information contained in $S$
were unavailable. Under these approximations, a larger $r(S)$ suggests
that the fixed policy relies more strongly on the corresponding visual
evidence for predicting the demonstrated action.

\paragraph{Connection to flow matching.}
For the flow-matching action head used in our experiments, the prediction
target $Y$ corresponds to the target velocity, and the action objective is
a squared regression loss. The setting in Proposition~1 therefore directly
matches the form of the flow-matching training objective.

When comparing the clean and intervened observations, we keep the
demonstrated action, robot state, language instruction, sampled flow time,
noise, noisy action, and target velocity unchanged. Thus, the two losses
differ only in the selected visual evidence, reducing stochastic variation
unrelated to the intervention.

\paragraph{Scope of the interpretation.}
The practical score $r(S)$ should not be interpreted as an unbiased estimate
of $\mathcal{V}(S)$ or as a physical causal effect. Visual suppression through
masking or blurring only approximates ideal information removal and may
introduce distribution shifts. In addition, the resulting relevance estimate
inherits the knowledge and biases of the initialized policy.

We therefore interpret $r(S)$ as a policy-level estimate of relative
predictive relevance. ActGaze uses these intervention responses as spatial
supervision for gaze learning, while the empirical improvements in visual
alignment and manipulation performance support their practical utility.

\subsection{Ground-Truth Region Annotation}
\label{app:gt_annotation}

\begin{figure}[t]
    \centering
    \includegraphics[width=\linewidth]{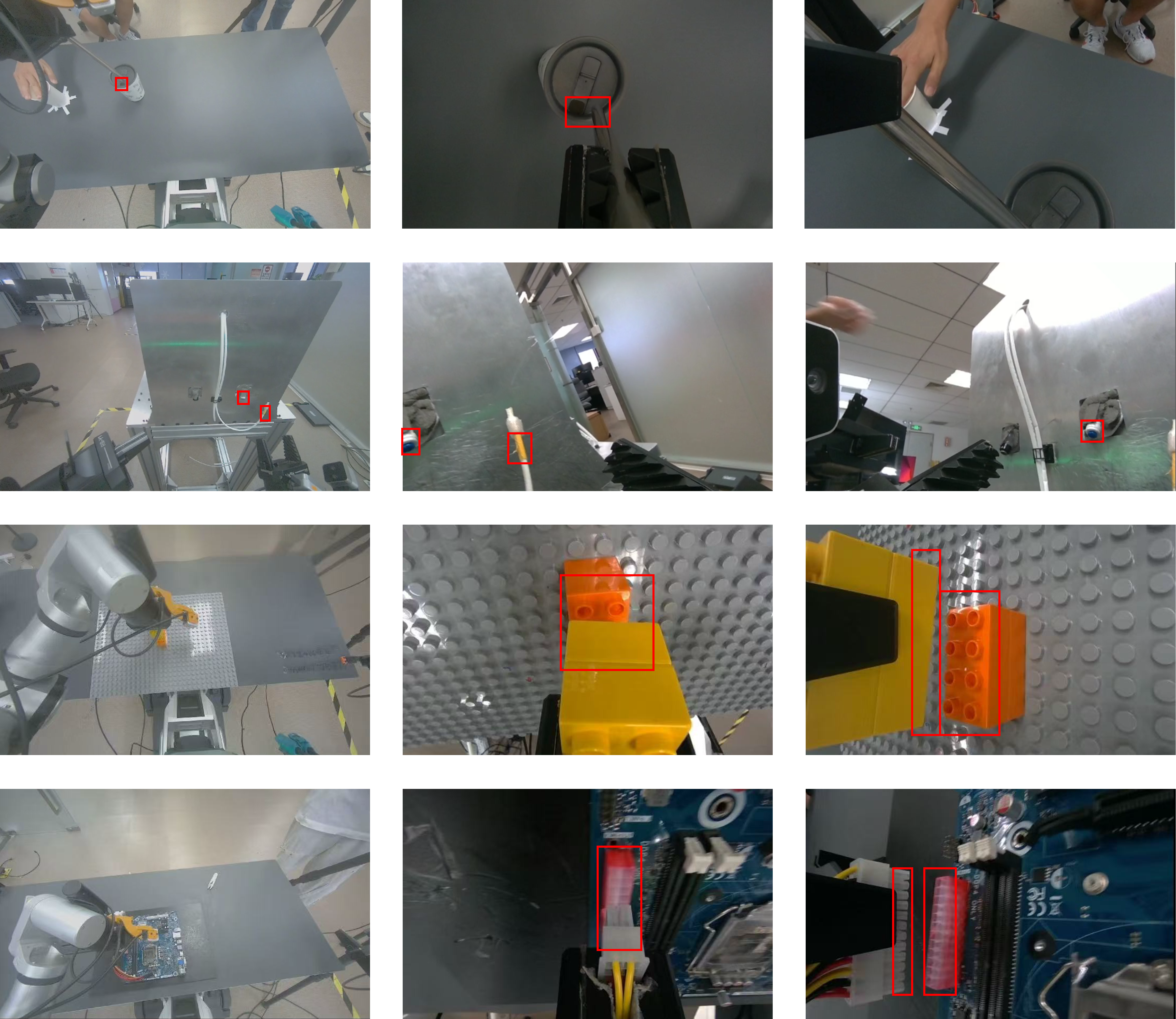}
    \caption{
    \textbf{Examples of Task-Relevant Region Annotations.}
    The annotations indicate localized visual regions containing evidence
    considered relevant to the current manipulation phase. The relevant
    region changes as the task progresses, e.g., from the manipulated object
    or grasping region during acquisition to the target opening or mating
    interface during alignment and insertion. Camera views that do not provide
    useful visual evidence for the current action are left unannotated.
    }
    \label{fig:gt_annotation_examples}
\end{figure}

The ground-truth task-relevant regions are used only for evaluating spatial
alignment and are never exposed to ActGaze during training. Representative
examples are shown in Fig.~\ref{fig:gt_annotation_examples}.

Four annotators participated in the annotation process following a shared
protocol. For each frame and camera view, annotators were instructed to mark
a localized region containing the visual evidence most relevant to the
current manipulation step, rather than broadly covering the entire object or
workspace.

The annotations are phase-dependent. During object acquisition, the relevant
region typically corresponds to the manipulated object or its local grasping
area. During alignment and insertion, the annotation shifts toward visual cues
that are more directly related to precise interaction, such as the target
opening, mating interface, or contact region. If a camera view does not contain
useful visual evidence for the current action, that view is left unannotated
instead of assigning an arbitrary region.

These annotations provide the reference regions used to compute the
Ground-Truth Alignment Score $S_{\mathrm{GT}}$ reported in the main paper.

\subsection{Task Specifications}
\label{app:tasks}

We evaluate ActGaze on four precision-demanding manipulation tasks:
Straw Insertion, Hose Insertion, LEGO Assembly, and Connector Plug-in.
Together, these tasks cover rigid insertion, deformable-object manipulation,
assembly, and multi-pin alignment. Their detailed specifications and
randomization settings are described below.

\begin{itemize}

    \item \textbf{Straw Insertion.}
    The robot grasps an $11.8$\,mm-wide straw and inserts it into a
    $13.8$\,mm-wide opening on a cup. The cup position is randomized within
    a $10$\,cm $\times$ $10$\,cm workspace to evaluate positional
    generalization.

    \item \textbf{Hose Insertion.}
    The robot grasps a $4.6$\,mm-wide compliant refrigerator drainage hose
    and inserts it into a $5.0$\,mm-wide opening. In addition to the tight
    insertion tolerance, deformation of the hose introduces further geometric
    variation during manipulation. The opening base is translated by up to
    $10$\,cm and rotated by up to $15^\circ$ from its nominal pose.

    \item \textbf{LEGO Assembly.}
    The robot grasps an eight-hole yellow LEGO brick and assembles it onto
    an orange mating brick. Each mating hole is $9.3$\,mm wide, and the
    tight fit requires accurate alignment between the two bricks. The yellow
    brick is randomized within a $5$\,cm $\times$ $5$\,cm workspace with
    rotations of up to $\pm30^\circ$, while the orange mating brick is
    randomized within a $15$\,cm $\times$ $15$\,cm workspace.

    \item \textbf{Connector Plug-in.}
    The robot grasps a 24-pin ATX connector and inserts it into the
    corresponding motherboard socket. The clearance between the connector
    and socket is within $0.4$\,mm, and simultaneous alignment of multiple
    pins makes the task particularly sensitive to small pose errors. The
    motherboard position is randomized by approximately $3$\,cm, while the
    connector position is randomized over a $10$\,cm horizontal range.

\end{itemize}
Overall, these tasks provide diverse test cases for evaluating ActGaze under varying geometric precision, object properties, and spatial randomization.

\subsection{Robustness to Visual Distractors}
\label{app:distractors}

We further evaluate whether the task-relevant gaze learned by ActGaze
improves robustness to visual perturbations. We conduct this study on LEGO
Assembly under three conditions: the original clean setting, additional
visual clutter, and strong lighting variation.

\begin{figure}[t]
    \centering
    \includegraphics[width=\linewidth]{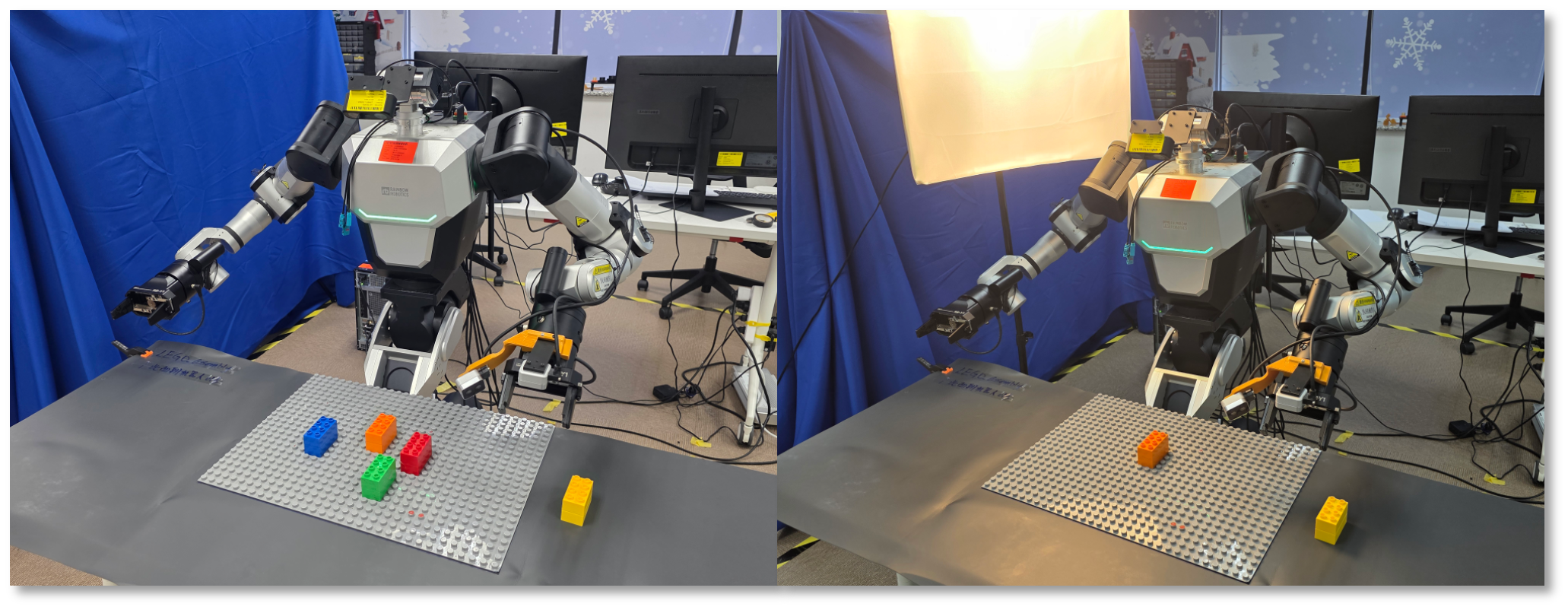}
    \caption{
    \textbf{Controlled Visual Variations.}
    Left: visual clutter introduced by additional LEGO pieces around the
    manipulation region. Right: strong lighting variation introduced by an
    external light source.
    }
    \label{fig:visual_distractors}
\end{figure}

For the visual-clutter condition, multiple LEGO pieces of different colors
are randomly placed around the target orange LEGO brick, introducing
task-irrelevant objects near the manipulation region. For the lighting
condition, a high-power external light source is introduced to produce a
substantial global appearance change. Representative examples of these two
conditions are shown in Fig.~\ref{fig:visual_distractors}.

\begin{table}[t]
\caption{
Robustness evaluation on LEGO Assembly under controlled visual variations.
}
\centering
\small
\begin{tabular}{lccc}
\toprule
Method & Clean & Clutter & Lighting \\
\midrule
$\pi_{0.5}$ & 18/20 & 13/20 & 11/20 \\
\textbf{ActGaze} & \textbf{19/20} & \textbf{15/20} & 11/20 \\
\bottomrule
\end{tabular}
\label{tab:visual_robustness}
\end{table}

As shown in Table~\ref{tab:visual_robustness}, both methods achieve similar
performance in the clean setting, while ActGaze improves the success rate
under visual clutter from 13/20 to 15/20. This result suggests that guiding
the policy toward task-relevant visual regions can reduce its sensitivity to
nearby task-irrelevant content.

Under strong lighting variation, both methods achieve 11/20, and no clear
improvement is observed with ActGaze. Unlike local visual clutter, lighting
variation affects the appearance of the entire observation, including the
task-relevant regions themselves. This result suggests that the robustness
benefit of ActGaze is more pronounced for spatial distractors than for global
appearance changes.